\documentclass{article}

\usepackage{arxiv}

\usepackage[utf8]{inputenc} 
\usepackage[T1]{fontenc}    
\usepackage{hyperref}       
\usepackage{url}            
\usepackage{booktabs}       
\usepackage{amsfonts}       
\usepackage{nicefrac}       
\usepackage{microtype}      
\usepackage{lipsum}
\usepackage{graphicx}
\usepackage{xcolor}
\usepackage{natbib}

\title{Declarative Machine Learning Systems}

\author{
 Piero Molino \\
  Department of Computer Science\\
  Stanford University\\
  \texttt{pmolino@cs.stanford.edu} \\

   \And
 Christopher Ré \\
  Department of Computer Science\\
  Stanford University\\
  \texttt{chrismre@cs.stanford.edu} \\
  
}

\begin{document}
\maketitle



\section{Introduction}

In the last twenty years machine learning (ML) has progressively moved from a academic endeavor to a pervasive technology adopted in almost every aspect of computing.
ML-powered products are now embedded in every aspect of our digital lives: from recommendations of what to watch, to divining our search intent, to powering virtual assistants in consumer and enterprise settings.
Moreover, recent successes in applying ML in natural sciences~\citep{Senior2020} had revealed that ML can be used to tackle some of the hardest real-world problems humanity faces today.
For these reasons ML has become central in the strategy of tech companies and has gathered even more attention from academia than ever before.
The process that led to the current ML-centric computing world was hastened by several factors, including hardware improvements that enabled the massively parallel processing
, data infrastructure improvements that enabled storage and consumption of massive datasets needed to train most ML models, and algorithmic improvements that allowed better performance and scaling.

Despite these successes, we argue that what we have witnessed in terms of ML adoption is only the tip of the iceberg.
Right now the people training and using ML models are typically experienced developers with years of study working within large organizations, but we believe the next wave of ML systems will allow a substantially larger amount of people, potentially without any coding skills, to perform the same tasks.
These new ML systems will not require users to fully understand all the details of how models are trained and utilized for obtaining predictions, 
a substantial barrier of entry,
but will provide them a more abstract interface that is less demanding and more familiar.
Declarative interfaces are well suited for this goal, by hiding complexity and favouring separation of interest, and ultimately leading to increased productivity.

We worked on such abstract interfaces by developing two declarative ML systems, Overton~\cite{Re2020} and Ludwig~\cite{Molino2019}, that require users to declare only their data schema (names and types of inputs) and tasks rather then writing low level ML code.
In this article our goal will be to describe how ML systems are currently structured, to highlight what factors are important for ML project success and which ones will determine wider ML adoption, what are the issues current ML systems are facing and how the systems we developed addressed them.
Finally we will describe what we believe can be learned from the trajectory of development of ML and systems throughout the years and how we believe the next generation of ML systems will look like.

\subsection{Software engineering meets ML}

A factor not enough appreciated in the successes of ML is an improved understanding of the process of producing real-world machine learning applications, and how different it is from traditional software development.
Building a working ML application requires a new set of abstractions and components, well characterized by \citet{Sculley2015}, who also identified how idiosyncratic aspects of ML projects may lead to a substantial increase in technical debt, i.e. the cost of reworking a solution that was obtained by cutting edges rather than following software engineering principles. 
These bespoke aspects of ML development are opposed to software engineering practices, with the main responsible being the amount of uncertainty at every step, which leads to a more service-oriented development process~\citep{casado_bornstein_2020}.

Despite the bespoke aspects of each individual ML project, researchers first and industry later distilled common patterns that abstract the most mechanical parts of the process of building ML projects in a set of tools, systems and platforms.
Consider for instance how the availability of projects like scikit-learn, TensorFlow, PyTorch, and many others, allowed for a wide ML adoption and quicker improvement of models through more standardized processes: where before implementing a ML model required years of work for highly skilled ML researchers, now the same can be accomplished in few lines of code that most developers would be able to write.
In a recent paper \citet{Hooker2020} argues that availability of accelerator hardware determines the success of ML algorithms potentially more than their intrinsic merits.
We agree with that assessment, and we add that availability of easy to use software packages tailored to ML algorithms has been at least as important for their success and adoption, if not more important.

\subsection{The coming wave of ML Systems}

\begin{figure}
    \centering
    \includegraphics[width=0.5\linewidth]{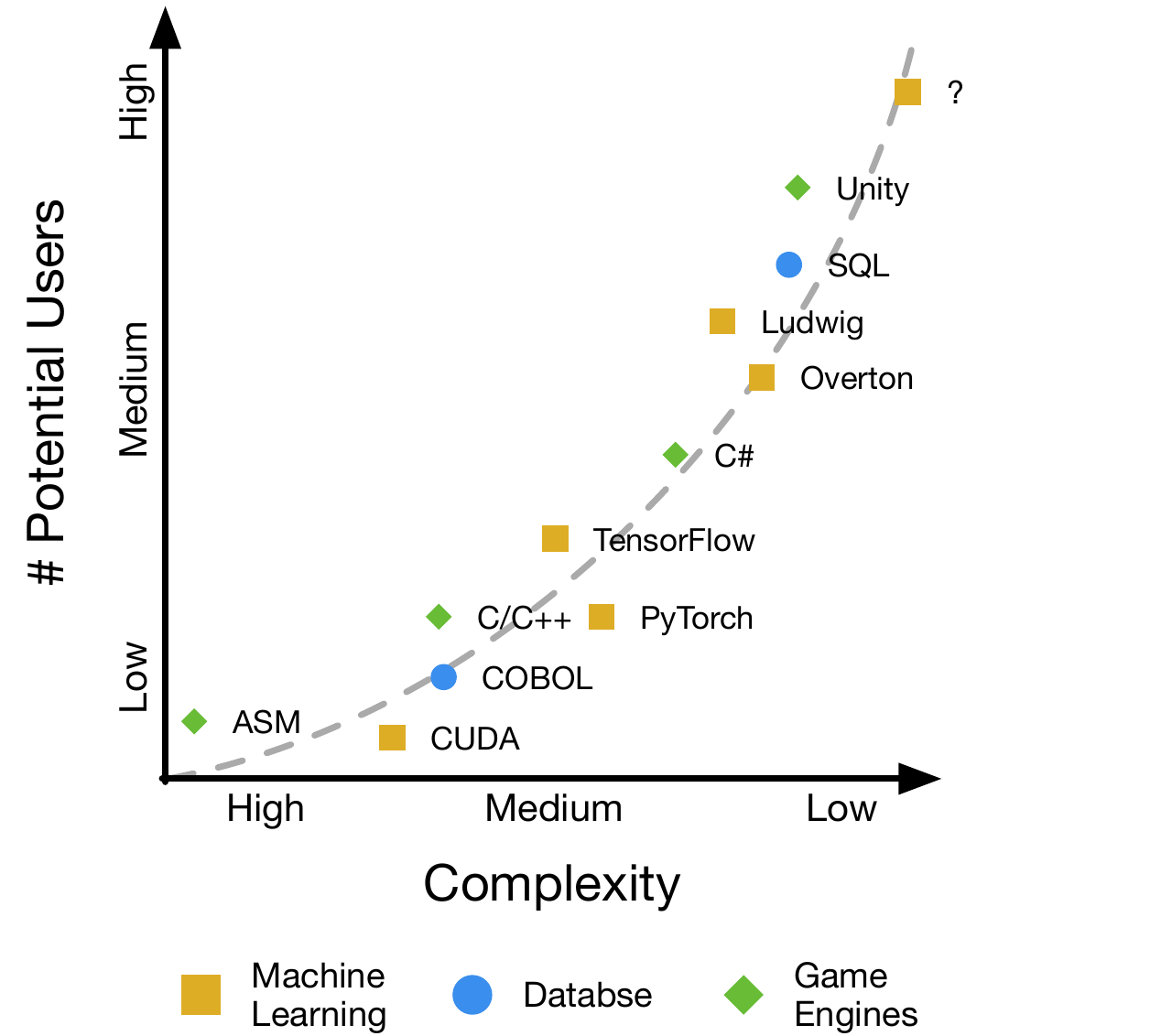}
    \caption{Approximate depiction of the relationship between how complex to learn and use a software tool is (either languages, libraries or entire products) and the amount of users potentially capable of using it, across different fields of computing.}
    \label{fig:complexity_vs_users}
\end{figure}

We believe generations of compiler, database, and operating systems work may inspire new foundational questions of how to build the next generation of ML-powered systems that will allow people without ML expertise to train models and obtain predictions through more abstract interfaces.
One of the many lessons learned from those systems throughout the history of computing is that substantial increases in adoption always come with separation of interests and hiding of complexity, as we depict in Figure~\ref{fig:complexity_vs_users}.
Like a compiler hides the complexity of low level machine code behind the facade of a higher level, more human-readable language, and as a database management system hides the complexity of data storage, indexing and retrieval behind the facade of a declarative query language, so we believe that the future of ML systems will steer towards  hiding complexity and exposing simpler abstractions, likely in a declarative way.
The separation of interests implied by such a shift will make it possible for highly skilled ML developers and researchers to work on improving the underlying models and infrastructure in a way that is similar to how today compiler maintainers and database developers improve their systems, while allowing a wider audience to use ML technologies by interfacing with them at a higher level of abstraction, like a programmer who writes code in a simpler language without knowing how it compiles in machine code or a data analyst who writes SQL queries without knowing the data structures used in the database indices or how a query planner works.
These analogies suggest that declarative interfaces are good candidates for the next wave of ML systems, with their hiding of complexity and separation of interest being the key to bring ML to non-coders.

We will provide a brief overview of the ML development life-cycle and the current state of ML platforms, together with what we identified as challenges and desiderata for ML systems.
We will also describe some initial attempts at building new declarative abstractions we worked on first hand that address those challenges.
We discovered that these declarative abstractions are useful for making ML more accessible to end users by avoiding having them write low-level error-prone ML code.
Finally, we will describe the lessons learned from these attempts and provide some speculations on what may lay ahead.

\section{Machine Learning Systems}


\subsection{Machine Learning development life-cycle}\label{sec:ml-dev-lifecycle}

\begin{figure}
    \centering
    \includegraphics[width=0.45\linewidth]{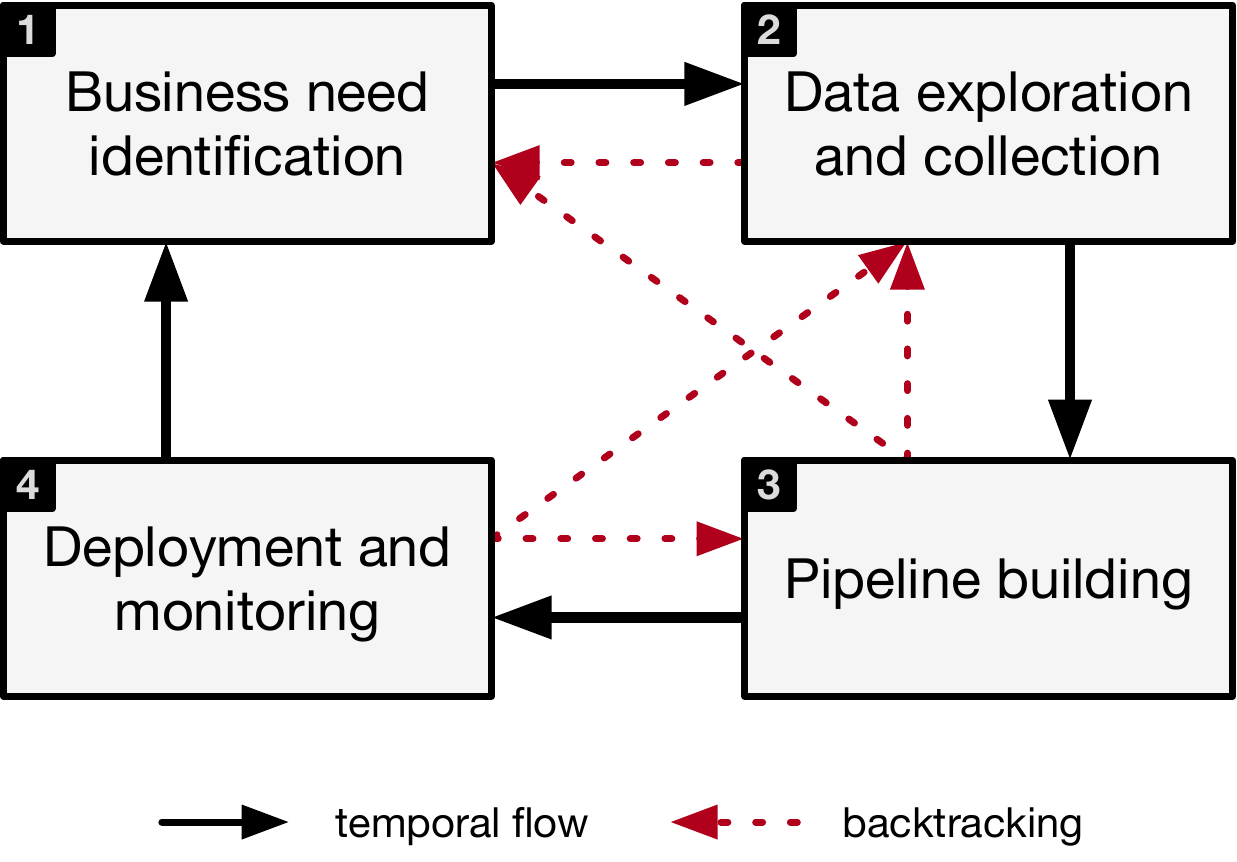}
    \caption{A four-steps coarse-grained model of the lifecycle of a ML project.}
    \label{fig:ml_lifecycle}
\end{figure}

Many descriptions of the development life-cycle of machine learning projects have been proposed, but the one we adopt in Figure~\ref{fig:ml_lifecycle} is a simple coarse-grained view that serves our exposition purposes, made of four high-level steps:

\begin{enumerate}
    \item business need identification
    \item data exploration and collection
    \item pipeline building
    \item deployment and monitoring
\end{enumerate}

Each of these steps is composed of many sub-steps and the whole process can be seen as a loop, with information gathered from the deployment and monitoring of an ML system helping identifying the business needs for the next cycle.
Despite the sequential nature of the process, each step's outcome is highly uncertain, and negative outcomes of each step may send the process back to previous ones.
For instance the data exploration may reveal that the available data does not contain enough signal to address the identified business need, or the pipeline building may reveal that, given the available data, no model can reach a high enough performance to address the business need and thus new data should be collected.

Because of the uncertainty that is intrinsic in ML processes, writing code for ML projects often leads to idiosyncratic practices: there is little to no code reuse and no logical / physical separation between abstractions, with even minor decisions taken early in the process impacting every aspect of the downstream code.
This is opposed to what happens, for instance, in database systems, where abstractions are usually very well defined: in a database system changes in how you store your data don't change your application code or how you write you queries, while in ML projects changes in the size or distribution of your data end up changing your application code, making code difficult to reuse.

\subsection{Machine Learning platforms and AutoML}\label{sec:ml-platforms-automl}

Each step of the ML process is currently supported by a set of tools and modules, with the potential advantage of making these complex systems more manageable and understandable.

Unfortunately, the lack of well defined standard interfaces between these tools and modules limits the benefits of a modularized approach and makes architecture design choices difficult to change.
The consequence is that these systems suffer from a cascade of compounding effects if errors or changes happen at any stage of the process, particularly early on, i.e. when a new version of a tokenizer is released with a bug, and all the following pieces (embedding layers, pretrained models, prediction modules) start to return wrong results.
The dataframe abstraction is so far the only widely used contact point between components, but it could be in itself problematic because of wide incompatibility among different implementations.

More end-to-end platforms
are being built mostly as monolithic internal tools in big companies to address the issue, but that often comes at the cost of having a bottleneck at either the organizational or at the technological level.
ML platform teams may become gatekeepers for ML progress throughout the organization, i.e. when a research team devises a new algorithm that does not fit into the mold of what the platform already supports, which makes productionizing the new algorithm extremely difficult.
These ML platforms can become a crystallization of outdated practices in the ever changing ML landscape.

At the same time AutoML systems promise to automate the human decision making involved in some parts of the process, in particular in the pipeline building, and, through techniques like hyperparameter optimization~\citep{Li2017}
and architecture search~\citep{Elsken2019}
, abstract the modeling part of the ML process.

AutoML is a promising direction, although research efforts are often centered around optimizing single steps of the pipeline (in particular finding the model architecture) rather than optimizing the whole pipeline, and the costs of finding a marginally better solution than a commonly accepted one may end up outweighting the gains.
In some instances this worsens the reusability issue, and contrasts with recent findings showing how architecture may actually not be the most impactful aspect of a model, as opposed to its size, at least for autoregressive models trained on big enough datasets~\citep{Henighan2020}.
Despite this, we believe that the automation AutoML brings is positive in general, as it allows developers to focus on what matters most and automate away more mundane and repetitive parts of the development process and help reduce the number of decisions they have to make.

We very much appreciate the intent of those platforms and AutoML systems to encapsulate best practices and simplify parts of the ML process, but we argue that there could be a better, less monolithic way, to think about ML platforms that may enable the advantages of these platforms while at the same time drastically reduce their issues, and that can incorporate the advantages of AutoML at the same time.

\subsection{Challenges and desiderata}\label{sec:challenges-and-desiderata}

Our experiences in developing both research and industrial ML projects and platforms led us to identify a set of challenges common to most of them and some desired solutions, which influenced us to develop declarative ML systems.

\begin{itemize}
    \item \textbf{Challenge 1: Exponential decision explosion.} Building a ML system involves many decisions, all need to be correct, with compounding errors at each stage. \textbf{Desideratum 1: Good defaults and automation.} Decisions should be reduced by nudging developers towards reasonable defaults and a repeatable automated process that makes those decisions (hyperparameter optimization for instance).
    \item \textbf{Challenge 2: ``New Model-itis''.} ML production teams try to build a new model and fail at improving performance for lack of understanding of the qulity and failure modes of previous models. \textbf{Desideratum 2: Standardization and focus on quality.} Low-added-value parts of the ML process should be automated with standardized evaluation and data processing and automated model building and comparison, shifting the attention from writing low level ML code to monitoring quality and improve supervision and shifting away the attention from mono-dimensional performance-based model leaderboards towards holistic evaluation.
    \item \textbf{Challenge 3: Organizational chasms.} There are gaps between teams working in pipelines that make it hard to share code and ideas, i.e. when entity disambiguation and intent classification teams are different in a virtual assistant project and don't share the codebase, which leads to replication and technical debt. \textbf{Desideratum 3: Common interfaces.} Increase reusability by coming up with standard interfaces that favor modularity and interchangeability of implementations.
    \item \textbf{Challenge 4: Scarcity of expertise.} Not many developers, even in large companies, can write low level Ml code. \textbf{Desideratum 4: Higher level abstractions.} Developers should not have to set hyperparameters manually or implement their custom model code unless really necessary, as it accounts for just a tiny fraction of the project lifecycle, and differences are usually tiny.
    \item \textbf{Challenge 5: Process slowness.} The process of developing ML projects can take months or years in some organizations to reach a desired quality because of the many iterations requires. \textbf{Desideratum 5: Rapid iteration.} Quality of ML projects improves by incorporating learnings from each iteration, so the faster each iteration is, the higher quality can be achieved in the same time. The combination of automation and higher level abstractions can improve the speed of iteration and in turn help improve quality.
    \item \textbf{Challenge 6: Many diverse stakeholders.} There are many stakeholders involved in the success of a ML project, with different skill sets and different interests, but only a tiny fraction of them has the capability to work hands-on on the system. \textbf{Desideratum 6: Separation of interests.} Enforcing a separation of interests with multiple user views would make a ML system accessible to more people in the stack allowing developers to focus on delivering value and improving the project outcome and consumers to tap into the created value more easily.
\end{itemize}

\section{Declarative ML Systems}

We believe that a declarative ML systems could fulfill the promise of addressing the above-mentioned challenges by implementing most of the desiderata.
The term may be overloaded in the vast literature of ML models and systems, so we restrict our definition of declarative ML systems to those systems that impose a separation between what a ML system should do and how it actually does it.
The ``what'' part can be declared with a configuration, that depending on its complexity and compositionality, can be seen as a declarative language, and can include information about the task to be solved by the ML system and the schema of the data it should be trained on.
This can be considered a low / no / zero code approach, as the declarative configuration is not an imperative language where the ``how'' is specified, so a user of a declarative ML system does not need to know how to implement a ML model or a ML pipeline as much as someone who writes a SQL query doesn't need to know about database indices and query planning.
The declarative configuration is translated / compiled into a trainable ML pipeline that respects the provided schema, and the trained pipeline can then be used for obtaining predictions.

Many declarative ML approaches have been proposed over the course of the years, most of which use either logic or probability theory or both as their main declarative interface.
Some examples of such approaches include probabilistic graphical models~\citep{koller2009} and their extensions to relational data such as probabilistic relational models~\citep{Friedman1999,Milch2005} and markov logic networks~\citep{Domingos2004} or purely logical representations such as Prolog and Datalog.
In these models domain knowledge can be specified as dependencies between variables (and relations) representing the structure of the model and their strength as free parameters.
Both the free parameters and the structure can be also learned from data.
These approaches were declarative in that they separate out the specification semantics from the inference algorithm.
However, performing inference on such models is in general hard and scalability becomes a major challenge.
Approches like Tuffy~\citep{Niu2011}, DeepDive~\citep{Zhang2017}, and others have been introduced to address the issue.
Nevertheless, by separating inference from representation, these models did a good job at allowing declaration of multitask and highly joint models, but were often outperformed by more powerful feature driven engines (e.g. deep learning based approaches).
We distinguish these declarative ML models from systems based on their scope: the latter focus on defining declaratively an entire production ML pipeline.

There are other potential higher level abstractions that hide the complexity of parts of the ML pipeline, and they have their own merits, but we do not consider them declarative ML systems.
Examples of such other abstractions could be libraries that allow users (ML developers) to write simpler ML code by removing the burden of having to write neural network layers implementations (like Keras
does) or having to write a for loop that is distributable and parallelizable (like PyTorch Lightning
does).
Other abstractions, like Caffe
, allow to write deep neural networks by declaring the layers of its architecture, but they do it at a level of granularity close to an imperative language.
Finally, abstractions like Thinc
provide a robust configuration system for parametrizing models, but also requires to write ML code that becomes parametrizable by the configuration system, thus not separating the ``what'' from the ``how''.

\subsection{Data-first}

Integrating data mining and ML tools has been the focus of several major research and industrial efforts since at least the 90s.
For example, Oracle's Dataminer shipped in 2001. 
They featured high-level SQL style syntax to use models and supported models defined externally in Java or via a standard called PMML.
These models were effectively syntax around user-defined functions to perform filtering or inference.
At the time, machine learning models were purpose-built using specialized solvers that required heavy-use of linear algebra packages, e.g., L-BFGS was one of the most popular for ML models.
However, the ML community began to realize that an extremely simple, classical algorithm called stochastic gradient descent (or incremental gradient methods) could be used to train many important ML models.
The Bismark project~\citep{Feng2012} showed that SGD could piggyback on existing data processing primitives that were already widely available in database systems (cf. with SciDB that rethought the entire database in terms of linear algebra).
In turn, integrating gradient descent and its variants allowed the DBMS to manage training.
This led to a new breed of systems that integrated training and inference.
They provided SQL syntax extensions to train models in a declarative way to manage training and deployment inside the database.
Examples of such systems are Bismark itself, MADLib~\citep{Hellerstein2012} (which was integrated in Impala, Oracle, Greenplum, etc.), and MLLib~\citep{Meng2016}.
The SQL extensions proposed in Bismark and MADlib are still popular as variants of this approach are integrated in the modeling language of Google's BigQuery~\citep{sato2012} or within modern open source systems like SQLFlow~\citep{wang2020}.

The data-centric viewpoint has the advantage of making models usable from withing the same environment where the data lives, avoiding potentially complicated data pipelines.
One issue that emerges is that, by exposing model training as a primitive in SQL, their users did not have fine-grained control of the modeling process.
For some class of models this became a substantial challenge, as the pain of piping the data to models (which these systems decreased substantially) was outweighed by the pain of performing  featurization and tuning the model.
As a result, many models lived outside the database.

\subsection{Models-first}

After successes of deep learning models in computer vision, speech recognition and natural language processing, the focus of both research and industry shifted towards a model-first approach, where the training process was more complicated and became the main focus.
A wrong implementation of backpropagation and differentiation would influence the performance of a ML project more than data preprocessing, and efficient computation of deep learning algorithms on accelerated hardware like GPUs transformed models that were too slow to train into the standard solution for certain ML problems, specifically perceptual ones.
In practice having an efficient wrapper of GPGPU libraries was more valuable than a generic data preprocessing pipeline.
Libraries like TensorFlow and PyTorch focused on abstracting the intricacies of low-level C code for tensor computation.

The availability of these libraries allowed for simpler model building, so researchers and practitioners started sharing their models and adapting others' models to their goals.
This process of transferring (pieces of) a pretrained model and tuning them on a new task started with word embeddings, but was later adopted also in computer vision, and now is made easier by libraries like Hugging Face's Transofrmers.

The two systems we built, Overton internally at Apple and Ludwig open source at Uber, are both model-first and focus on modern deep learning models, but also retrieve some features of the data-first approach, specifically the declarative nature, by adding separation of interest, and are both capable to use transfer learning.

\subsubsection{Overton in a nutshell}

Overton~\citep{Re2020} spawned from the same observations expressed at the beginning of this article: commodity tools changed the landscape of ML to the point that we can build tools capable of moving developers up the stack and allow users to focus and quality and quantity of supervision.
It was designed to make sure that people did not need to write new models for production applications in search, information extraction, question answering, named entity disambiguation, and other tasks while making it easy to evaluate models and improve performance by ingesting additional relevant data to get quality results on end deployed models.

\begin{figure}
    \centering
    \includegraphics[width=0.8\linewidth]{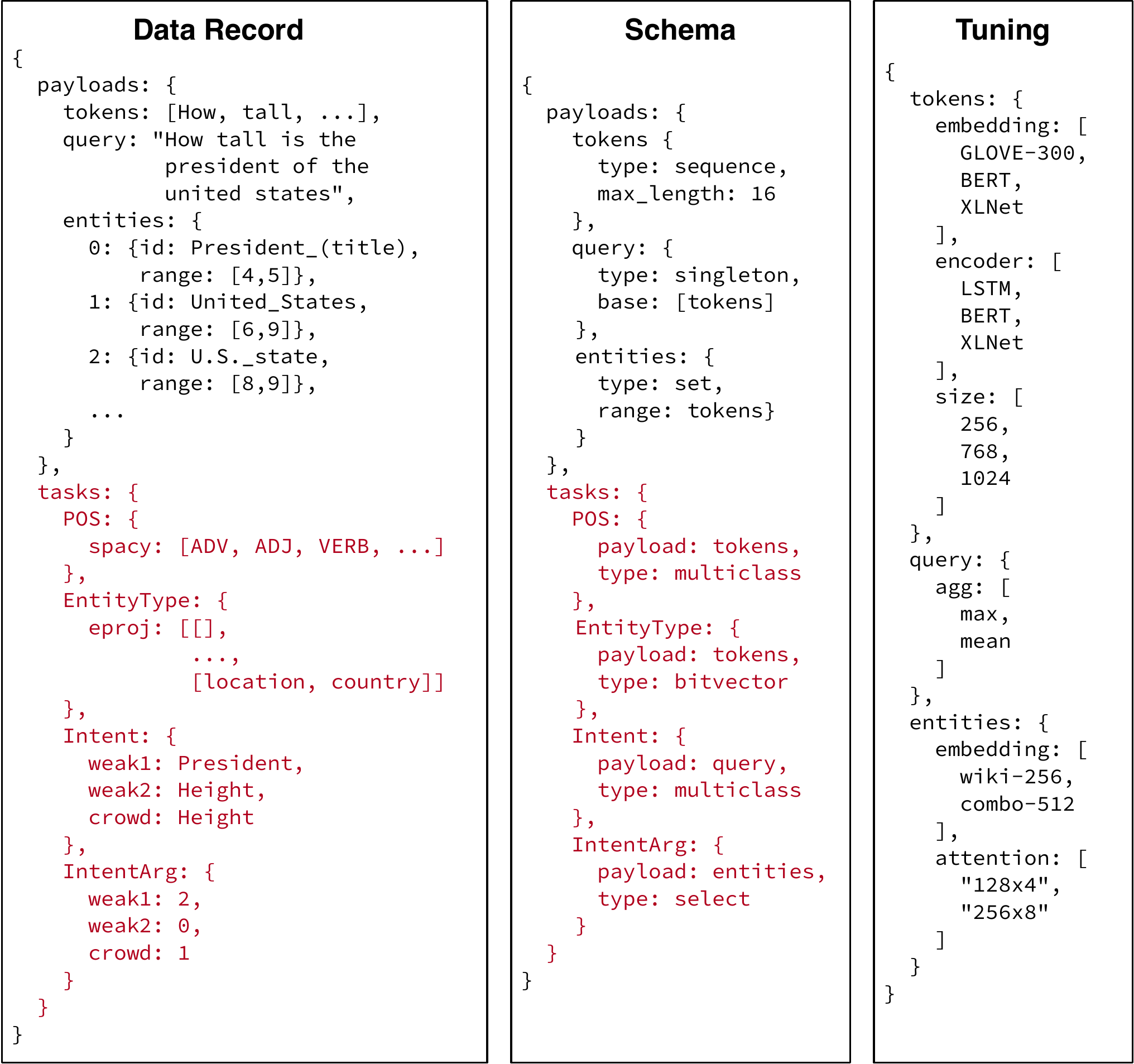}
    \caption{Example of an Overton application to a complex NLP task. On the left we show an example data record of a piece of text, with its payload (inputs, query, tokenization and candidate entities) and tasks (outputs parts of speech, entity type, intent and intent arguments), In the middle we show the Overton schema, detailing both payloads for the inputs and tasks for the outputs, with their respective types and parameters. On the right we show a tuning specification that details the coarse-grained architecture options Overton will choose from and compare for each payload.}
    \label{fig:overton_examples}
\end{figure}

Inspired by relational databases, a user would declare a schema that describes the incoming data source called payload.
In addition, a user would also describe a high-level data flow among the tasks, with optionally multiple sources of (weak) supervision, as shown in Figure~\ref{fig:overton_examples}.
The system is able to use this barebones information to compile trainable models (including data preprocessing and symbols mappings), combine supervision using data programming techniques~\citep{Ratner2016}, compile a model in TensorFlow, PyTorch or CoreML, and produce performance reports and finally export a deployable model in CoreML, TensorFlow or ONNX.

A key technical idea was that many sub-problems like architecture search or hyperparameter optimization could be done with simple methods, like coarse grained architecture search (only classes of architectures are chosen, not all their internal hyperparameters) or very simple grid search. 
A user could override some of these decisions, but custom options were not heavily optimized in runtime. 
Other features included multi-task learning, data slicing and the use of pretrained models.

The role of Overton users becomes monitoring performance, improve supervision quality by adding new examples and providing new forms of supervision, they don't need to write models in low level ML code.
Overton was responsible for massive gains in quality (40\% to 82\% error reduction) in search and question answering applications at Apple and, as a consequence, the footprint of the engineering team was substantially reduced and no one was writing low level ML code.

\subsubsection{Ludwig in a nutshell}

Ludwig~\citep{Molino2019} is a system that allows its users to build end-to-end deep learning pipelines through a declarative configuration, train them and use them for obtaining predictions.
The pipelines include data preprocessing that transforms raw data into tensors, the model architecture building, the training loop, the prediction, the postprocessing of data and the evaluation of pipelines.
Ludwig also includes a visualization module for model performance analysis and comparison, and a declarative hyperparameter optimization module.

One key idea of Ludwig is that it abstracts both the data schema and the tasks as data types feature interfaces so that users only need to define a list of input features and outputs features, both with their names and data type.
This allows for modularity and extensibility: the same text preprocessing code and the same text encoding architecture code is reused every time a model that includes text features is instantiated, while the same multi-label classification code for prediction and evaluation is adopted every time set feature is specified as an output, for instance.

This flexibility and abstraction is made possible by the fact that Ludwig is opinionated about the structure of the deep learning models it builds, following the encoders-combiner-decoders (ECD) architecture introduced by \citet{Molino2019}, which allows for easily defining multi-modal and multi-task models, depending on the data type of both inputs and outputs available in the training data.
The ECD architecture also defines precise interfaces, which greatly improve code reuse and extensibility: by imposing the dimensions of the input and output tensors of an image encoder, for instance, the architecture allows for many interchangeable implementations of image encoding (i.e. a CNN stack, or a stack of residual blocks, or stack of transformer layers) and choosing which one to use in the configuration requires just changing one string parameter.

\begin{figure}
    \centering
    \includegraphics[width=0.7\linewidth]{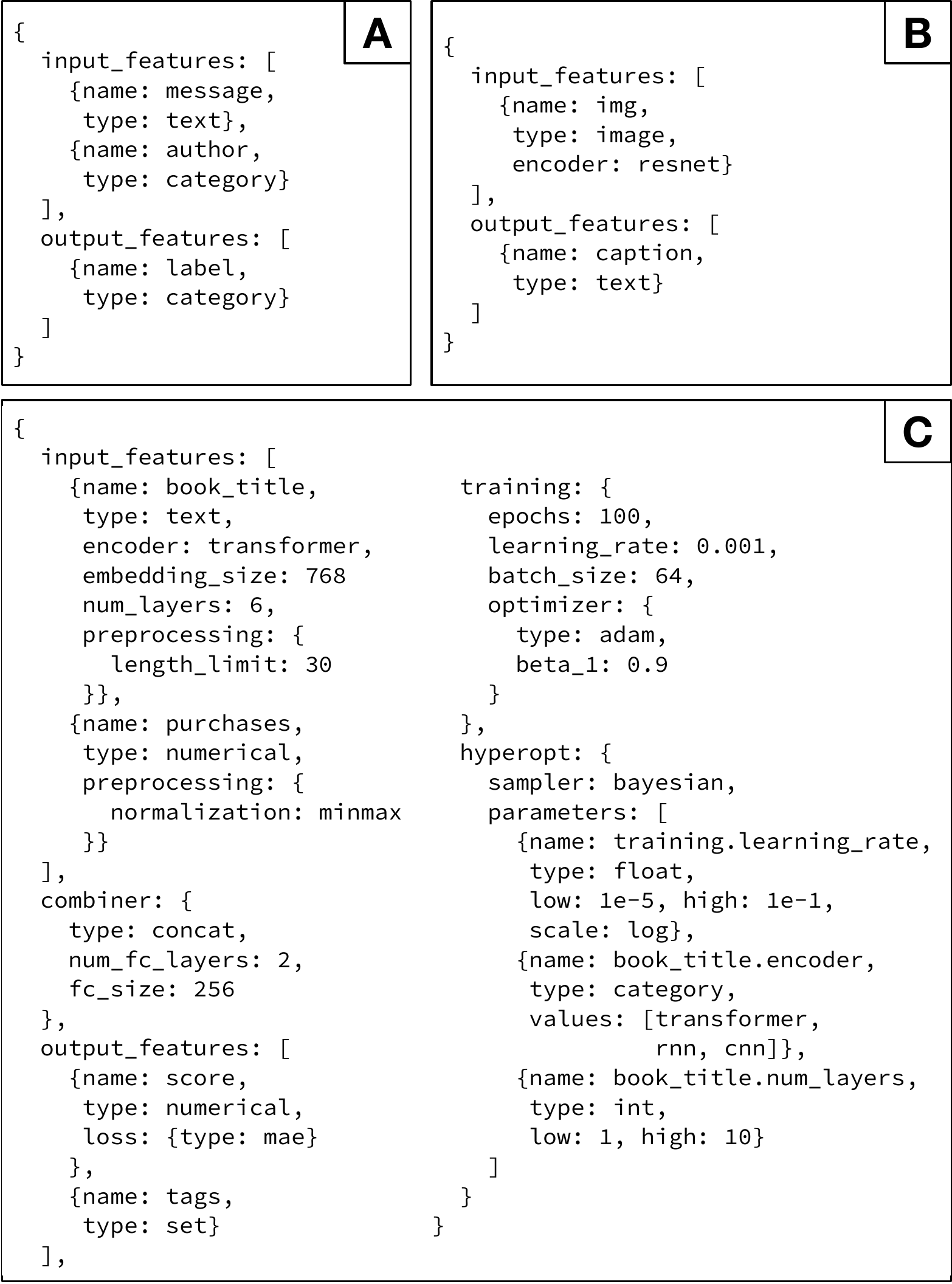}
    \caption{Three examples of Ludwig configurations. A: a simple text classifier that includes additional structured information about the author of the classified message. B: image captioning example. C: A detailed configuration for a model that given the title and sales figures of a book, predicts it's user score and tags. A and B show simple configurations, while C shows the degree of control of each encoder, combiner and decoder, together with training and preprocessing parameters, while also highlighting how Ludwig supports hyperparameter optimization of every possible configuration parameter.}
    \label{fig:ludwig_examples}
\end{figure}

What makes Ludwig general is that, depending on the combination of types of inputs and outputs declared in the configuration, the specific model instantiated from the ECD architecture solves a different task: input text and output category will make Ludwig compile a text classification architecture, while an image input and a text output will result in an image captioning system, and both image and text inputs with a text output will result in a visual question answering model.
Moreover, basic Ludwig configurations are very easy to write and hide most of the complexity of building a deep learning model, but at the same time they allow the user to specify all details of the architecture, the training loop and the preprocessing if they so desire, as shown in Figure~\ref{fig:ludwig_examples}.
The declarative hyperopt section shown in Figure~\ref{fig:ludwig_examples}(C) makes it possible to automate architectural, training and preprocessing decisions.

In the end, Ludwig is both a modular and an end-to-end system: the internals are highly modular for allowing Ludwig developers to add options, improve the existing ones and reuse code, but from the perspective of the Ludwig user, it's entirely end-to-end (including processing, training, hyperopt, and evaluation).

\subsubsection{Similarities and differences}

Both Overton and Ludwig, despite being developed entirely independently of each other, converged on similar design decisions, in particular on the adoption of declarative configurations that include (albeit with a different syntax) both the input data schema and a notion of the tasks models should solve in Overton and the analogous notion of input and output features in Ludwig.
Both systems have a notion of types associated with the data which inform parts of the pipeline they build.

Where the two systems differ is in some assumptions, some capabilities and in what they focus on.
Overton is more concerned with being able to compile its models in various different formats, in particular for deployments, while Ludwig has only one productionization route.
Overton also allows for a more explicit way to define data-related aspects like weak supervision and data slicing.
Ludwig on the other hand covers a a wider breadth of use cases by virtue of the compositionality of the ECD architecture, where different combinations of inputs and outputs can define different ML tasks.

Despite the differences, both systems address some of the challenges highlighted in Section~\ref{sec:challenges-and-desiderata}.
Both systems nudge developers towards making less decisions by automating part of the life-cycle (Desideratum 1) and towards reusing models already available to them and analyze them thoroughly by providing both standard implementation of architectures and evaluations that can also be combined in a more holistic way (Desideratum 2).
The interfaces and the use of data types and associated higher level abstractions (Desideratum 4) in both systems favor code reuse (Desideratum 3) and addresses the expertise scarcity.
Declarative configurations increase the speed of model iteration (Desideratum 5) as developers just need to change details in the declaration instead of rewriting code with cascade effects.
Both systems also partially provide separation of interests (Desideratum 6): they separate between the system developers adding new models, types and features and the users using the declarative interface.

\section{What is next?}

The adoption of both Overton and Ludwig in real-world scenarios by tech companies suggests that they are actually solving at least part of the concrete problems those companies face and produce value.
Despite that, we believe there is substantial more value to be untapped by combining their strengths with the tighter integration with data of the ``data-first'' era of declarative ML systems.
This new wave of recent deep learning work has shown that with relatively simple building blocks and AutoML, fine-control of the training and tuning process may no longer be necessary, thus solving the main pain point that ``data-first'' approaches did not address and opening the door for a convergence towards new, higher level systems that seamlessly integrate models training, inference and data.

In this regard, there are lessons that can be learned from computing history, by observing the process that led to the emergence of general systems that replaced bespoke solutions:
\begin{itemize}
    \item \textbf{Amount of users}. We believe even higher level abstractions are needed for ML to become not only more widely adopted, but to end up being developed, trained, improved and used by people without any coding skill. We believe that, to draw another analogy with database systems, we are still in the ``Cobol'' era of ML, and that as ``SQL'' allowed the a substantially larger amount of people to write database application code,
    the same will happen for ML.
    \item \textbf{Explicit user roles}. We expect not everyone interacting with a future ML system to be trained in ML, statistics, or even computer science. Similarly to how databases evolved to the point that there's a stark separation between database developers implementing faster algorithms, database admins managing instances installation and configuration and database users writing application code and final users obtaining fast answers to their requests, we expect this role separation to emerge in ML systems.
    \item \textbf{Performance optimizations}. More abstract systems tend to make compromises either in terms of expressivity or in terms of performance. Ludwig achieving state of the art and Overton replacing production systems suggest that may be a false tradeoff already. Compilers history suggests a similar pattern: over time optimized compilers could often beat hand-tuned machine code kernels, although the complexity of the task may have suggested otherwise initially. We believe developments in this directions will make it so that bespoke solutions would likely be limited to really specific tasks in the fat part of the (growing) long tail of ML tasks within an organization, where even very minor improvements are extremely valuable, similarly to the mission critical use cases where today one may want to write assembly code.
    \item \textbf{Symbiotic relationship between systems and libraries}. We believe the will be more ML libraries in the future and that they will co-exist and help improving ML systems in virtuous cycle. In the history of computing this has happened over and over, with a recent example being the emergence of full-text indexing libraries such as Lucene filling the feature gap most DBMSs had at the time, with Lucene being used in bespoke applications first and later being used as the foundation for complete search systems like ElasticSearch and Solr and being finally integrated in DBMSs like OrientDB, GraphDB and others.
\end{itemize}

Some challenges are still open for declarative ML systems: they will have to demonstrate to be robust with respect to future changes in machine learning coming from research, supporting diverse training regimes and showing that the types of task they can represent encompasses a large fraction of practical uses.
The jury is still out on this.

To conclude, we believe that technologies change the world when they can be harnessed by more people than the ones who can build them, so we believe the future of machine learning and how impactful it will be in everyone's life ultimately depends on the effort of putting it in the hands of the rest of us.

\section*{Acknowledgements}
The authors want to thank Antonio Vergari, Karan Goel, Sahaana Suri, Chip Huyen, Dan Fu, Arun Kumar and Michael Cafarella for insightful comments and suggestions.

\bibliographystyle{plainnat}  
\bibliography{references}

\end{document}